\documentclass[runningheads,a4paper]{llncs}

\usepackage{amssymb}
\usepackage{amsmath}
\usepackage{graphicx}

\usepackage{url}
\urldef{\mailsa}\path|{ilkka.huopaniemi, tommi.suvitaival, samuel.kaski}@tkk.fi|
\urldef{\mailsb}\path|}@tkk.fi|
\urldef{\mailsc}\path|matej.oresic@vtt.fi|    
\newcommand{\keywords}[1]{\par\addvspace\baselineskip
\noindent\keywordname\enspace\ignorespaces#1}

\begin{document}

\mainmatter  % start of an individual contribution

\title{Translating biomarkers between multi-way time-series experiments from multiple species}

\titlerunning{Translating biomarkers between multi-way time-series experiments}

\author{Ilkka Huopaniemi$^{1}$%
\thanks{I.H., T.S and S.K belong to the Adaptive Informatics Research Centre. The work was funded by Tekes MASI program and by Tekes Multibio project. I.H. is funded by the Graduate School of Computer Science
and Engineering. S.K is partially supported by EU FP7 NoE PASCAL2, ICT
216886.}
\and Tommi Suvitaival$^{1}$ \and  Matej Ore\v{s}i\v{c}$^{2}$ \and Samuel Kaski$^{1*}$\\}
\authorrunning{Ilkka Huopaniemi et al.}

\institute{1 Aalto University School of Science and Technology, Department of Information and Computer Science, Helsinki Institute for Information Technology HIIT, P.O. Box 15400, FI-00076 Aalto, Finland\\
2 VTT Technical Research Centre of Finland, P.O. Box 1000, FIN-02044 VTT, Espoo, Finland \\
\mailsa\\
\mailsc\\
\url{http://www.cis.hut.fi/projects/mi}}
\toctitle{Lecture Notes in Computer Science}
\tocauthor{Authors' Instructions}
\maketitle
\begin{abstract}
  Translating potential disease biomarkers between multi-species
  'omics' experiments is a new direction in biomedical research. The
  existing methods are limited to simple experimental setups such as
  basic healthy-diseased comparisons. Most of these methods also
  require an \emph{a priori} matching of the variables (e.g., genes
  or metabolites) between the species. However, many experiments have
  a complicated multi-way experimental design often involving
  irregularly-sampled time-series measurements, and for instance
  metabolites do not always have known matchings between organisms. We
  introduce a Bayesian modelling framework for translating between
  multiple species the results from 'omics' experiments having a
  complex multi-way, time-series experimental design. The underlying
  assumption is that the unknown matching can be inferred from the
  response of the variables to multiple covariates including time.
\end{abstract}
\section{Introduction}
Cross-species analysis of biological data is an increasingly important
direction in biological research. The analysis calls for multivariate methods, since 'omics' technologies, such as transcriptomics
and metabolomics, enable studying the dynamic response of biological
organisms in various conditions, including various time points during disease progression. An important reserach problem of
translational medicine is translating potential biomarkers for disease
between species. This would ultimately allow mapping phenotypes
between model organisms and actual human experiments.

The basic experimental design in searching for disease biomarkers is the one-way
comparison of healthy and diseased populations. At the simplest,
biomarkers can be translated between species by comparing lists of $p$-values of simple
differential expression. Most existing cross-species analysis methods are limited to such simple
designs \cite{Lu09}. A further limitation of most existing methods is that they require an \emph{a priori}  matching of
variables (genes) between the organisms. Such orthology information is
not always available, especially in metabolomics where the
mapping of metabolites between organisms has barely
started, and is an interesting research problem in itself.

Most biological experiments have a multi-way experimental design,
where healthy and diseased populations are further divided into
subpopulations according to additional covariates such as gender,
treatment groups, age, measurement times etc. A usual approach for
dealing with the additional covariates is stratifying the
diseased-healthy comparison; a typical example is comparing healthy
and diseased males and females separately. The
standard statistical methods for properly dealing with multi-way
designs, are Analysis of Variance (ANOVA) and its multivariate
generalization (MANOVA). While studying the effects of all the
covariates on the data makes the analysis slightly more complicated,
more information can be gained from each species to be used in the
translation. Unfortunately, there exist no earlier proper tools for utilizing
the information of the effects of multiple covariates on the data in
cross-species analysis.

Furthermore, time-series experiments are becoming more and more common
in clinical studies searching for disease biomarkers. Whereas in
some cases the measurement times of such experiments are regular and
allow a ``neat and easy'' data analysis, this is often not the
case. In clinical follow-up studies, such as \cite{Oresic08},
measurement times are often irregular due to practical reasons of data
collection, and there are missing time points. Also, in follow-up
studies spanning timescales of years, individuals have been shown to
develop into metabolic developmental states at an individual pace
\cite{Nikkila08}. In addition, life spans of different organisms, such
as man and mouse, are very different resulting in very different
measurement times. These complications call for a possibility to align
the time-series. All of these factors combined cause remarkable
challenges for cross-species data analysis.

Instead of searching for single molecule biomarkers, which have a high
risk of false positives, we concentrate on finding combinations of
similarly-behaving biomarkers, which is a way towards treating
a transcriptomic or a metabolic profile as a fingerprint of the
clinical status of the organism. For this, multivariate statistics is
needed. In this paper, we will now show how the data analysis problem
can be formulated as a new multivariate ANOVA-type model in the case where data
comes from multiple sources (species) and one of the variables, namely the time,
has a previously unknown structure (alignment).

In this paper, we will present a formal framework for cross-species
analysis of 'omics' data in the case of a multi-way, time-series
experimental design. This methodology can be directly used for finding
previously unknown matching of groups of variables between the
species based on the data. In contrast to many-step approaches, the whole modelling
is done in a single, unified, multivariate Bayesian model. The
framework has estimation of uncertainty and dimensionality
reduction built-in to overcome the main challenge: high dimensionality and
small sample-size. The central underlying assumption is that the actual link between the variables of the different species can be inferred from a similar response of the
variables to multi-way covariates.
\subsection*{Previous work in cross-species analysis}
Until now, meta-analysis of microarray data has been the major
approach to cross-species studies in biology \cite{Lu09}. Plenty of
meta-analyses have been done by either comparing lists of
differentially expressed genes between several species, or by
comparing expression levels of known gene orthologs between the
species. So far, mainly highly-controlled cell cycle studies have been
analyzed across several species and no attention has been paid to
multi-way designs.

One step towards translational analysis has been taken by Lucas
\textit{et al.} \cite{Lucas09sagmb,Lucas09}. To help finding biomarkers
from an \textit{in vivo} experiment, they incorporated prior knowledge
from results of an \textit{in vitro} study analyzed with the same
method. Like ours, their approach is based on generative Bayesian modeling of
factors and can handle multiple covariates. Their approach does not,
however, consider time-series cases with unaligned time nor the case
without any \emph{a priori} matching between the variables. 

A probabilistic model based on Gaussian random fields has been applied
to two-species expression data \cite{Lu10jcb}. This work combined
 differential expression scores from different species, cell
types, and pathogens utilizing homology information. %The authors were
%able to extract a core set of innate immune response genes that are
%active over the covariates. The uncertainty in the found set is not explicit
%in the results and a high number of variables compared to
%samples may pose a problem in the analysis, as the covariate-specific
%profile is estimated for each variable separately.
In \cite{Le10} the task was to query large databases of micro-array
experiments to identify similar experiments in different species, by
utilizing partially known orthology information. In \cite{Liu10csb}
time-series micro-array data from multiple species was used to
discover causal relations between genes to discover conserved
regulatory networks. Also this approach naturally needs \emph{a priori} known
matching of orthologous genes.

A standard method for finding similarities between several data sets
is canonical correlation analysis (CCA)\cite{Hotelling36}. CCA assumes
paired samples over the data set and thus is not directly applicable
for the translation problem, where the samples (patients) are
different over species. A simple iterative method for pairing genes
has been developed in \cite{Tripathi09icassp,Tripathi10dmkd}. In the case the genes are samples in the data matrix, optimal pairing of genes is sought by maximizing the dependency between the data
sets estimated by CCA. A very similar method was recently used for
regulatory network inference \cite{Gholami10}. No prior matching of variables or samples is assumed, and the method attempts to find both
iteratively by alternating between matching of variables and matching of samples
using a closely related method Co-Inertia analysis. These methods do
not, however, take into account covariate information (including
measurement time) of the samples, nor different time resolution of the covariates.

In summary: none of the existing approaches can take into account a
multi-way time-series covariate structure or exploit it to find
previously unknown matchings between the variables without any \emph{a priori} known matching information.
%\bibliographystyle{splncs03}
%\bibliography{/share/mi/mi/doc/mi,/share/mi/mi/doc/gene/gene}
%\end{document}
\section{Model}
Our method addresses the problem of translating biomarkers between
multiple species from multi-way time-series experiments with
previously unknown matchings between the variables (metabolites). We do
not assume the samples have a pairing but our main assumption is that there is a
similar multi-way time-series experimental design in both experiments.

A simple data analysis procedure towards this goal would be doing the univariate
multi-way ANOVA analysis separately on the two data sets of the two
species, and comparing the lists of $p$-values afterwards as a meta-analysis
step.

In the two-way case, to explain the covariate-related variation in one species, say ${\bf x}$, the following linear model is
usually assumed:
\begin{equation}
    {\bf x}_j|_{(a,b)}=\boldsymbol{\mu}^x+\boldsymbol{\alpha}^x_a+\boldsymbol{\beta}^x_b+(\boldsymbol{\alpha\beta})^x_{ab}+\epsilon_j.
\label{ANOVAmodel}
\end{equation}
Here ${\bf x}_j$ is a continuous-valued data vector, observation
number $j$, the $\boldsymbol{\mu}^x$ is the overall (grand) mean, the $a$ and $b$ ($a=0,\dots A$ and $b=0,\dots B$) are
the two independent covariates, such as disease and treatment. The
$\boldsymbol{\alpha}^x_a$ and $\boldsymbol{\beta}^x_b$ are parameter vectors
describing the covariate-specific effects, called main effects. The
$(\boldsymbol{\alpha\beta})^x_{ab}$ is a parameter vector describing the interaction effect. 

Instead of searching for single-molecule biomarkers, that have a high
risk of false positives, our approach is multivariate, concentrating
on finding combinations of biomarkers.

In order to tackle high dimensionality and scarcity of observations,
we assume that there are groups of similarly behaving variables
(metabolites) in each species. We then search for correlated groups of
metabolites sharing a similar response to external covariates.  These
correlated groups (clusters) are therefore assumed to be shared
between the species. Underlying this process is the assumption that
similarity of multi-way behavior of groups of metabolites indicates a
cross-species mapping of the metabolites.

We have taken an ambitious goal by building a unified Bayesian model
that integrates the separate multi-way experiments from multiple
species. The model can be learned jointly by Gibbs sampling.

From the point of view of ANOVA-type modelling the question
is how to do multi-way modelling when the data comes from different sources
with different variables ({\it e.g.}, man and mouse having different
metabolites). The solution is to consider data ``source'' as an additional
covariate in the multi-way analysis \cite{Huopaniemi10ecml}.  From the
data integration point of view the task is to find dependencies
between the data sets when neither the variables nor the samples have been
paired. There is, however, a shared multi-way covariate structure in
the data sets, and it is utilized to find the mapping of groups of
variables. We study additionally the case where one of the covariates,
time, has a previously unknown structure due to unknown time
alignments. It will be shown that the alignments can be found
simultaneously within the whole unified model.

The model we develop for this task is an extension of our recently
published multi-way modelling methods
\cite{Huopaniemi09,Huopaniemi10ismb,Huopaniemi10ecml}. In
\cite{Huopaniemi09}, we presented a method for multi-way ANOVA-type
modelling in ``small sample-size $n$, high dimensionality $p$
''-conditions in the case of standard covariates, such as disease,
treatment, and gender. The solution is to use regularized factor
analysis for dimensionality reduction, such that each variable is
assumed to come from one factor only. The effects of multi-way
covariates $\boldsymbol{ \alpha}_a,\boldsymbol{ \beta}_b,(\boldsymbol{ \alpha\beta})_{ab}$ are then estimated in the low-dimensional latent factor
space. Each latent factor represents a group of correlated variables. The model is 
\begin{eqnarray}
  {\bf x}_{j}^{lat} \sim \mathcal{N}(\boldsymbol{ \alpha}_a+
  \boldsymbol{ \beta}_b+(\boldsymbol{ \alpha\beta})_{ab},\mathbf{I}) \nonumber \\
  {\bf x}_j\sim \mathcal{N}({\boldsymbol{\mu}}+{\bf V} {\bf
    x}_{j}^{lat},{\mathbf{\Lambda}}) \; . \label{FA_model}
\end{eqnarray}
Here ${\bf x}_{j}$ is a $p$-dimensional data vector, $\mathbf{V}$ is
the projection matrix, and ${\bf x}_{j}^{lat}$ is the low-dimensional latent variable,
${\mathbf{\Lambda}}$ is a diagonal residual variance matrix with
diagonal elements $\sigma_i^2$. 

In \cite{Huopaniemi10ismb}, we further extended this framework into
integrating data sources with {\it paired samples}, such as having
measurements from multiple tissues of each individual. In
\cite{Huopaniemi10ecml}, we first extended \cite{Huopaniemi09} into
time-series cases with unknown alignments, such that these alignments
can be learned simultaneously with the multi-way modelling task. In
\cite{Huopaniemi10ecml}, we also presented the basic principle and a
simplified model of how the multi-way modelling framework can be
extended into translational modelling. This case is much more
difficult than \cite{Huopaniemi10ismb}, because samples have not been paired
between the data sources; For example, the pairing of one time-point
of an individual test mouse and a time-point from one of the human
patients cannot be assumed. In \cite{Huopaniemi10ecml} we concentrated on finding a shared response of
the variables to one covariate only; the aligned
time. In this paper we now proceed by presenting the full
translational model where, in addition to aligned time, there are
other covariates, such as disease. Also, in this paper we separate the
time- and disease behavior into shared and species-specific effects.
\subsection{Modelling time-series measurements from multiple populations with regular measurement times}
Let us now consider modelling data from time-series measurements from
diseased and healthy populations in one species. If the measurement
times are fixed and individuals can be assumed to have similar aging
development, the data analysis can be seen as a two-way design and
modelled with a linear model. When modelling the effects of time and
other covariates on low-dimensional latent factors, each factor
representing one correlated group of variables \cite{Huopaniemi09}, we can use the model
\begin{equation} {\bf x}_j^{lat}|_{(t,b)} =\boldsymbol{ \alpha}_t+
  \boldsymbol{ \beta}_b+(\boldsymbol{ \alpha\beta})_{tb}+\textrm{noise}.
\end{equation}
We denote from now on the time-point by $t$, the disease status by
$b=\{0,1\}$, the effect of time by $\boldsymbol{\alpha}_t$, the effect
of disease by $\boldsymbol{ \beta}_b$, and the interaction of time and
disease $(\boldsymbol{ \alpha\beta})_{tb}$. The last one is the most
interesting, denoting the time-dependent disease effects.
\subsection{Modelling time-series measurements from multiple populations with irregular measurement times}
This work is motivated by the fact that in many real-world
sparsely collected time-series datasets, especially from large-cohort human clinical studies, measurement times
can be irregular within and between individuals; one particular
state-of-the-art clinical lipidomic study is \cite{Oresic08}, on which we now
concentrate. This study followed a set of patients; some of them remained healthy, some developed into type 1
Diabetes. Furthermore, it was shown in \cite{Nikkila08} that
individuals progress into different age-related metabolic states at their
indivual pace. This phenomenon can be modelled by assuming that there are
underlying latent metabolomic development states and individuals
progress into these states in their individual pace \cite{Nikkila08}. The underlying states were modelled by Hidden Markov
Models (HMM), where the observed metabolic profiles are assumed to be emitted by
the underlying states. This modelling assumption also deals with the
problem of aligning irregular measurements.

The important problem now is how to separate the effects of disease from
the individual aging changes.  In \cite{Nikkila08} the HMM model was
trained separately for the healthy population and the diseased
population, and such an approach cannot fully answer this question.

The model \cite{Huopaniemi10ecml} that can separate these two effects is 
\begin{equation}
\mathbf{x}^{lat}_j|_{state(j,t)=s, b }\sim \mathcal{N}(\boldsymbol {\alpha}_s
+\boldsymbol {\beta}_b+(\boldsymbol{\alpha \beta})_{sb},\mathbf{I}),
\label{HMM-time}
\end{equation}
where $s$ is the latent development state (HMM-state), $\boldsymbol{\alpha}_s$, is the effect of aligned HMM-time and $(\boldsymbol{\alpha
  \beta})_{sb}$ is the most interesting effect, the interaction of
``HMM-time'' and disease. We showed in \cite{Huopaniemi10ecml} that it
is possible to simultaneously estimate the terms in the model
(\ref{HMM-time}) and learn the alignments of the time-series into the
HMM development states. We assume a linear HMM-chain, allowing only self-transitions and transitions into the next state. The probability of the $t$:th time-point of individual $j$ being in state $s$ is
\begin{equation}
p(s(j,t)=s)=p \left(s(j,t)|s(j,t-1) \right)p \left( x_j^{lat}|\boldsymbol {\alpha}_s
+\boldsymbol {\beta}_b+(\boldsymbol{\alpha \beta})_{sb} \right)p \left(s(j,t+1)|s(j,t) \right).
\end{equation}
If more covariates are present in the study, it is straightforward to
extend the model (\ref{HMM-time}) by additional terms.
\subsection{Translating biomarkers between species from time-series
  measurements from multiple populations with irregular measurement
  times}
\begin{figure}
 \centering
 \begin{tabular}{l|ll}
 a) & \hspace{0.001\textwidth} & b)\\
 \includegraphics[width=0.6\textwidth]{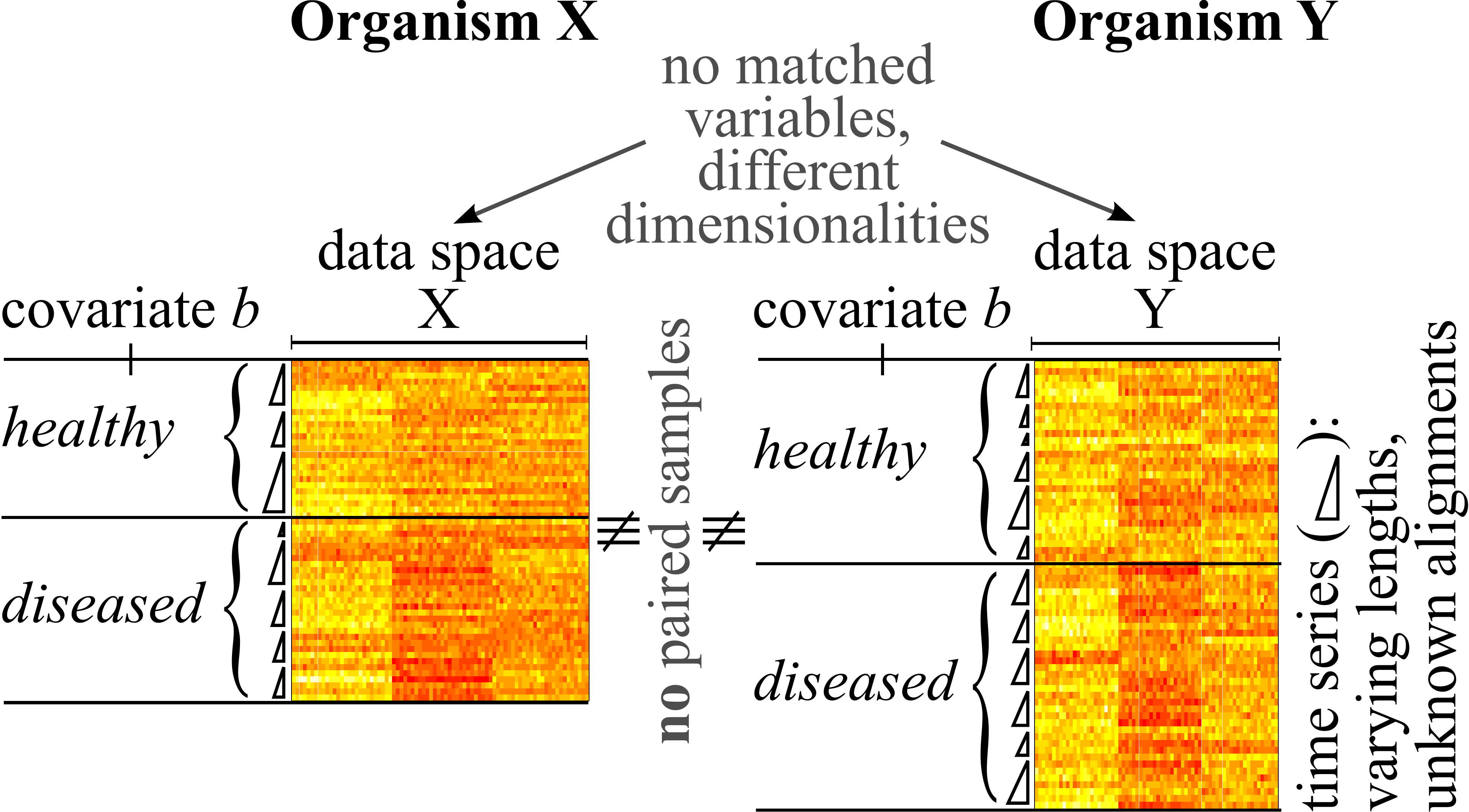} \hspace{0.001\textwidth} & &
\includegraphics[width=0.37\textwidth]{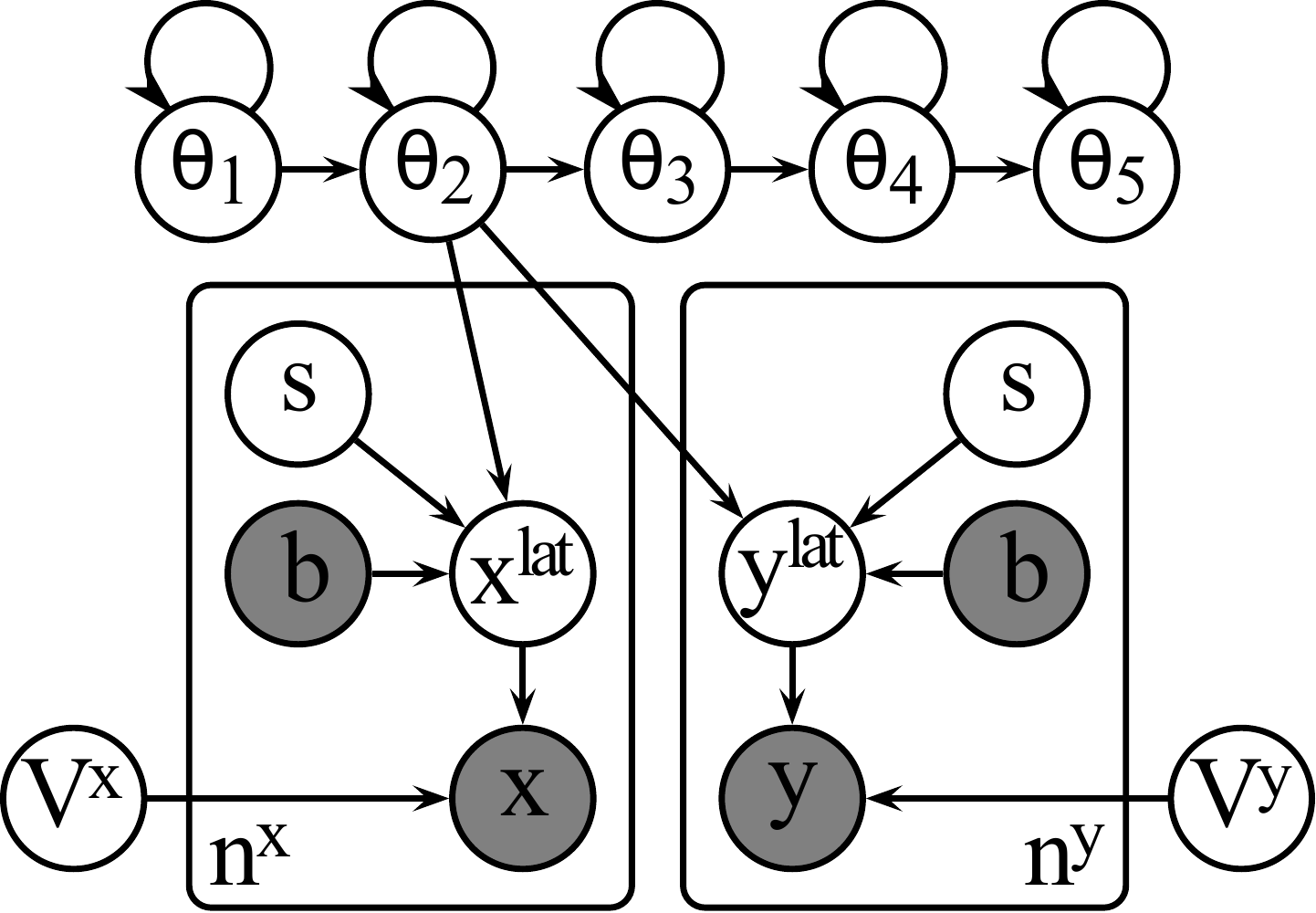}
 \end{tabular}
 \caption{a) Data matrix representation of the data analysis problem. (b) Plate diagram of the Bayesian graphical model. The set $\boldsymbol{\theta}_s = \{ \boldsymbol{\alpha}_s, \boldsymbol{\alpha}^x_s, \boldsymbol{\alpha}^y_s, (\boldsymbol{\alpha\beta})_{sb}, (\boldsymbol{\alpha\beta})^x_{sb}, (\boldsymbol{\alpha\beta})^y_{sb} \}$ contains all latent variables describing the corresponding HMM state. The state of each sample is determined by an observed covariate $b$ and an unobserved covariate $s$.}
 \label{fig:Platediagram}
\end{figure}

We now propose that translation of results between multiple species, from
multi-way time-series experiments, should be done by finding groups of
similarly behaving variables (metabolites) in both species that
respond similarly into multi-way covariates. A data matrix representation of the data analysis problem and plate diagram of the Bayesian model are shown in Figure ~\ref{fig:Platediagram}.
We introduce a modeling
framework that can do this task, even in the complicated case of having
irregular time-series measurements that require alignment into hidden
metabolic states.

The model makes the very flexible assumption \cite{Huopaniemi10ecml} that the observed data
vectors in the two species with different variables, {\bf x} and {\bf y}, are generated by
latent effects according to
\begin{eqnarray}
  {\bf x}=\boldsymbol{\mu}^x+f^x(\boldsymbol{\alpha}_s+\boldsymbol {\beta}_b+(\boldsymbol{\alpha \beta})_{sb})+f^x(\boldsymbol{\alpha}^x_s+\boldsymbol {\beta}^x_b+(\boldsymbol{\alpha \beta})^x_{sb})+\epsilon,\nonumber \\
  {\bf
    y}=\boldsymbol{\mu}^y+f^y(\boldsymbol{\alpha}_s+\boldsymbol {\beta}_b+(\boldsymbol{\alpha \beta})_{sb})+f^y(\boldsymbol{\alpha}^y_s+\boldsymbol {\beta}^y_b+(\boldsymbol{\alpha \beta})^y_{sb})+
  \epsilon \; ,
\label{3wANOVAmodelf_nopairing}
\end{eqnarray}
where $\boldsymbol{\alpha}_s$, $\boldsymbol {\beta}_b$,
$(\boldsymbol{\alpha \beta})_{sb}$ are the shared effects of HMM-time,
disease and interaction of HMM-time and disease, respectively,
$\boldsymbol{\alpha}^x_s$, $\boldsymbol {\beta}^x_b$ and
$(\boldsymbol{\alpha \beta})^x_{sb}$ and are the species $×$-specific
effects of HMM-time, disease and interaction of HMM-time and disease,
respectively, and likewise for species $y$. The variable-spaces of
${\bf x}$ and ${\bf y}$ are different, and therefore also the
dimensions of the latent variables of $\mathbf{x}^{lat}_j$ and
$\mathbf{y}^{lat}_j$ representing groups of correlated variables in
both species, need not match. For this reason, the latent effects of the covariates have to be projected into the actual observed data spaces ${\bf x}$ and ${\bf y}$ through previously unknown projections $f^x$ and $f^y$, that will be learned jointly with the model. 

The translational problem now becomes: Does some dimension of
$\mathbf{x}^{lat}_j$ respond to the covariates $s$ and $b$ similarly
as one of $\mathbf{y}^{lat}_j$. If it does, one can represent this
behavior with {\bf shared} effects $(\boldsymbol{\alpha}_s,
\boldsymbol {\beta}_b,(\boldsymbol{\alpha \beta})_{sb})$. The
interpretation is that a cluster of correlated variables in ${\bf x}$
represented by the dimension of $\mathbf{x}^{lat}_j$ matches with a
cluster of correlated variables in ${\bf y}$. Such dimensions can be
considered as multi-species biomarkers. If there is no match, the
response to the external covariates is modelled by species-specific
effects ($\boldsymbol{\alpha}^x_s,\boldsymbol
{\beta}^x_b,(\boldsymbol{\alpha \beta})^x_{sb}) $). With this
framework, we are able to estimate confidence of the shared
effects. 

\subsection{Matching problem}

We propose the following measure for quantifying the quality of the
match between two clusters from different datasets: whether the
matching is better than an average matching (over other pairs). On a
meta-level the measure is intuitively appealing in the spirit of
permutation tests, and it can be formulated more exactly by specifying
what we mean by ``better.'' We will use probabilistic modeling to
measure relative goodness below.

The matching problem of the clusters is a combinatorial problem, where
possible configurations of pairs need to be evaluated, judging for
each pair how similarly they respond to multi-way covariates. We
resort to an iterative algorithm that attempts to change the matching
of one cluster at a time. Choosing a candidate pair, we compare its
goodness to an average pair (uniformly selected having one same
endpoint), and accept forming a link between them by a Metropolis
criterion that compares the likelihoods of the two pairings. A reverse
operation is to attempt to break a link by comparing an existing link
between two clusters to an average (random) pair. The goodness
(likelihood) of a pair is evaluated by a shared multi-way model
between the clusters. Clusters with no pairs are modelled as specific
effects. Averaging over the iterations, we can estimate the
probability for matchings and the ``shapes'' of the multi-way
effects. A high probability of a specific pair indicates a found
matching. A high probability for being modelled as a specific effect
indicates the cluster has no pair.

\section{Results}
We illustrate the method with generated data and lipidomic time series
data with a two-way, time-series experimental design. In the
experiments, we neglect the static disease effects
$\boldsymbol{\beta_b}$ and assume all the disease effects are due to
HMM-state-specific disease effects $(\boldsymbol{\alpha}
\boldsymbol{\beta})_{sb}$.
\subsection{Generated data}
We generated from the model two data sets $X$ and $Y$ with no pairing
of samples but only a shared two-way design. There are 11 separate
time-series (``patients'') in both of the two data matrices, each series
consisting of 5 to 15 time points. This results in 108 and 115 samples,
and data matrices are 200- and 210-dimensional. The latent factors
$\mathbf{x}^{\text{lat}}_j$ and $\mathbf{y}^{\text{lat}}_j$ are 3- and
4-dimensional, respectively. The data in each population is generated
from a shared HMM-chain with 5 states. We generate three covariate
effects into the data: (i) a shared temporal effect
$\boldsymbol{\alpha}_s$ as $0$, $+0.5$, $+1$, $+1.5$, $+2$ in one cluster
of data set X and one cluster of data set Y, (ii) a shared interaction effect
$(\boldsymbol{\alpha\beta})_{sb}$ as $0$, $-0.5$, $-1$, $-1.5$, $-2$ in another cluster of X and another cluster of Y, (iii) a specific temporal
effect $\boldsymbol{\alpha}^y_s$ as $0$, $-0.5$, $-1$, $-1.5$, $-2$ in
yet another cluster of data set Y. Patterns (i) and (ii) are the
only behavior that is shared between the two data sets representing
the two species, and the model should be able to learn the correct
pairing of variable clusters based on this similarity.
\begin{figure}
  \centering
%  \begin{tabular}{l||l}
%  \begin{tabular}{ll}
%  a) & Main effects $\alpha$:\\
%  & \textit{HMM-time}
%  \end{tabular}
%  &
%  \begin{tabular}{ll}
%  b) & Interaction effects $(\alpha\beta)$:\\
%  & \textit{Interaction of HMM-time and disease}
%  \end{tabular}\\\\
%  \includegraphics[width=0.49\textwidth]{figureGen.pdf} &
%  \includegraphics[width=0.49\textwidth]{figureGen-interaction.pdf}
%  \end{tabular}
 \includegraphics[width=\textwidth]{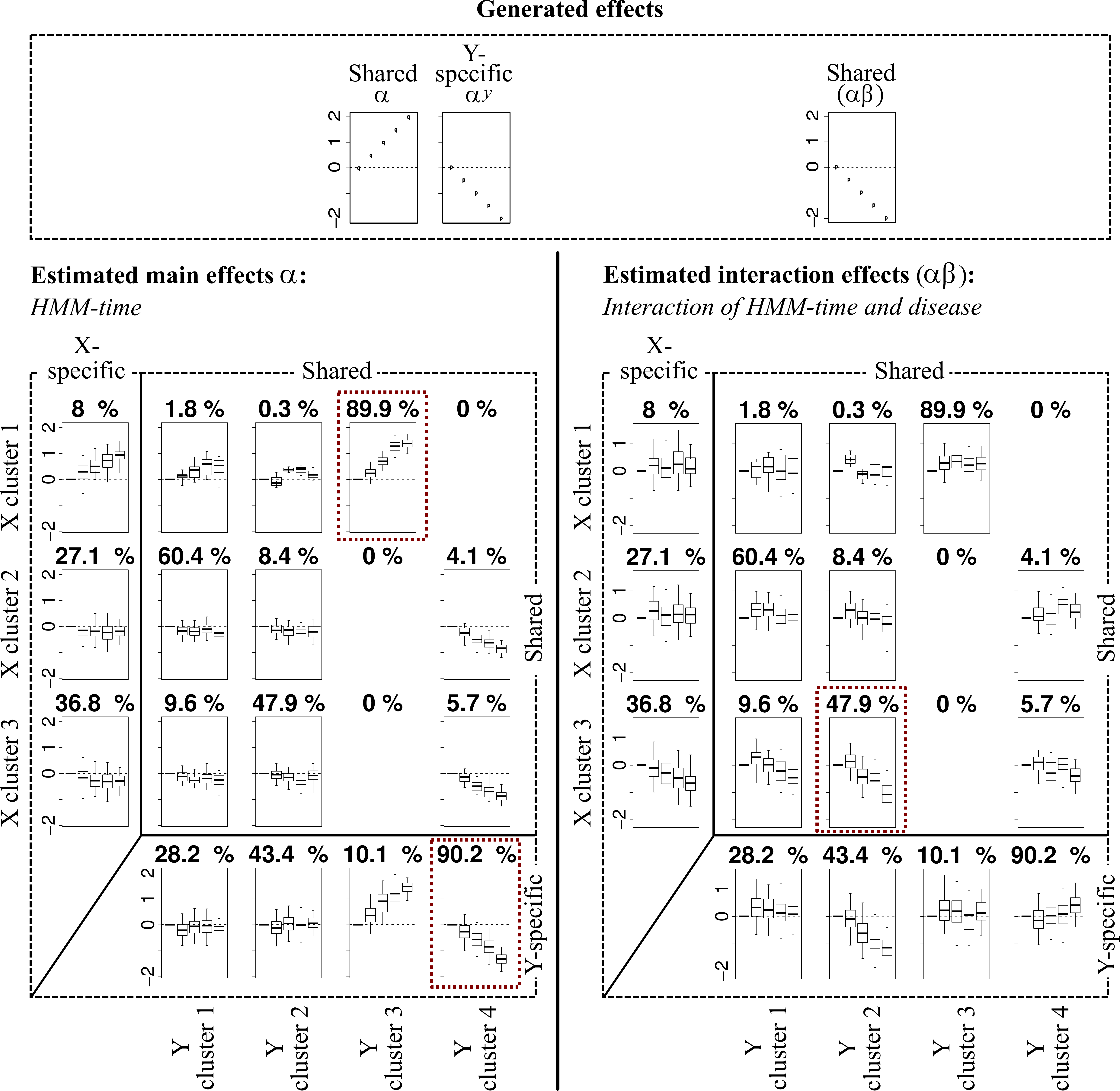}
 \caption{Pairing results from generated time-series
  data. Shown are the main effects (HMM-time; left) and interaction effects (right). Topmost, the generated effects are illustrated. Lower, the table of estimated effects
  shows shared (top-right area) and specific (left column and bottom
  row) effects for both types of effects. Rows and columns in the area of
  shared effects correspond to clusters in data sets X and Y,
  respectively. The found true pairing is highlighted by a red box. Value on top of each plot shows the percentage of
  posterior samples where the effect is found. An effect above or below zero is considered significant.}
\label{fig:figureGen}
\end{figure}
We use the proposed model to jointly align the samples into HMM
states, learn the clusters of variables, search for the possible
pairing of the clusters between the two data sets, and model the
ANOVA-type effects acting on the found clusters. We chose \emph{a priori}
a model with 5 HMM states. For the analysis we discarded the
first 5~000 samples from the Gibbs sampler as a burn-in period and
after this ran the model for another 5~000 Gibbs samples of which we
saved every 5th sample to obtain a total of 1~000 Gibbs samples to
approximate the underlying posterior distribution.

Our model finds the previously generated clusters without mistakes (see Fig.~\ref{fig:figureGen}). It also connects the X-cluster 1 to Y-cluster 3 and X-cluster 3 to Y-cluster 2 by estimating them to be linked in 89.9~\% and 47.9~\% of posterior samples, respectively. The generated shared and specific effects are detected as expected and no false positive effects are found.
\subsection{Lipidomic time-series data}
We then validate the model with a real lipidomic data set, consisting
of time-series measurements from a recently published type 1 diabetes
(T1D) follow-up study \cite{Oresic08}. In the data, there are 71
healthy patients and 53 patients that later developed into T1D. For
each time-series, there are 3--29 time points from irregular
intervals. For this validation study, we randomly divide the
individuals into two non-overlapping data
sets X and Y. We then study, whether we can find a similar response to
the covariates disease and HMM-time, and a matching between
those clusters from the two data sets that respond to these covariates.

Again, we \emph{a priori} chose a model with 5 HMM states. We
learned a two-way model for the data, where the first way is
``HMM-time'' and the second way is ``disease''. Lipids were assigned
into 6 clusters. We discarded 10~000 burn-in samples and collected
10~000 Gibbs samples of which we saved every 10th sample to obtain a
total of 1~000 Gibbs samples approximating the posterior distribution.

\begin{figure}
 \centering
 \includegraphics[width=\textwidth]{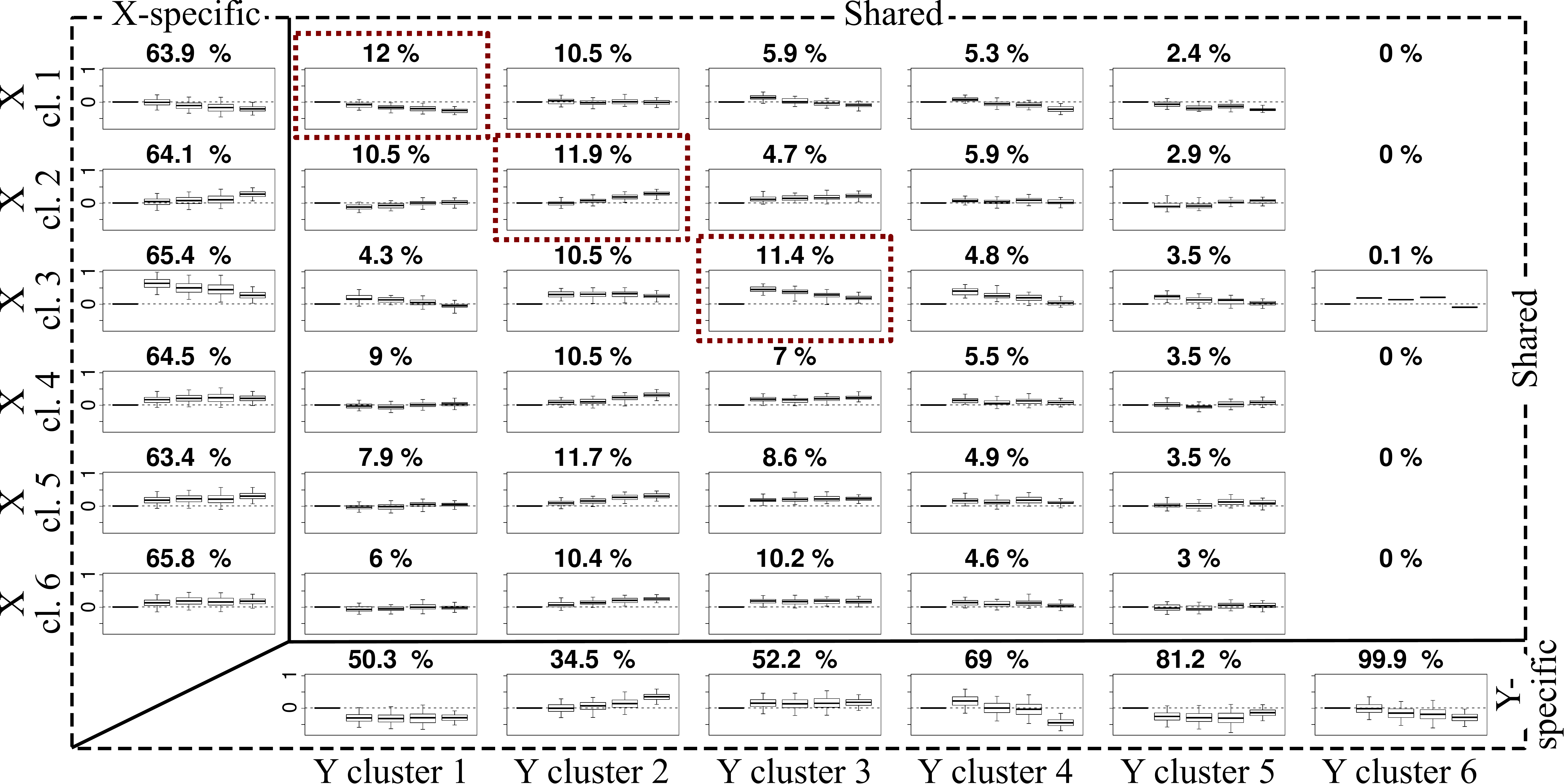}
 % figureDIPP.pdf: 1260x1260 pixel, 72dpi, 44.45x44.45 cm, bb=
 \caption{Pairing results from lipidomic time-series
   data. Only the main effects $\boldsymbol{\alpha}_s$ are shown.
   The true pairings found are highlighted by red boxes. The table shows
   shared (top-right area) and specific (left column and bottom row)
   effects estimated by the model. Rows and columns in the area of
   shared effects correspond to clusters in data sets X and Y,
   respectively. An effect (boxplot) consistently above or below zero is considered significant. Value on top of each plot shows the percentage of
   iterations where the clusters were matched.}
 \label{fig:figureDIPP}
\end{figure}

The model integrates the datasets X and Y by learning the HMM-time effects
$\boldsymbol{\alpha}_s$ and interaction effects $(\boldsymbol{\alpha
  \beta})_{sb}$. Clusters of lipids and effects found (Fig. ~\ref{fig:figureDIPP}) were similar as in our previous publication \cite{Huopaniemi10ecml}. 
The model finds three matching clusters between X and Y responding similarly to the external covariates, thus
linking the same lipids between the two subsets of data without prior
knowledge. The corresponding clusters were paired in 10--12~\% of
posterior samples, which is higher than for other combinations of
pairing (0--11~\%). Naturally the method does not find matchings for clusters that do not respond to external covariates. 

A group of triglycerol (TAG) and two
groups of glycerophosphocholine (GpCho) were strongly paired to their
counterparts. % (see Table~\ref{table:DIPPclusters}).
On real data, the result is naturally not as good as on generated data, since the effects are weaker.
%  \begin{table}
%    \caption{Lipids found in both clusters of X and Y, when the X-Y cluster pairs are according to the maximum \textit{a posteriori} (MAP; corresponding percentage of posterior samples shown) estimate of the pairing. Shown are three cluster pairs with the highest posterior probability.}
%    {\begin{tabular}{|l|}
%    %    \hline
%        \textbf{Cluster 1} TG(42:0) TAG(16:0/14:0/16:0) TAG(18:1/16:0/12:0) TG(46:2) \\ TAG(16:0/16:0/16:0) TAG(18:1/14:0/16:0) TAG(16:0/18:2/14:1) TAG(16:0/18:1/15:0) \\TAG(16:0/18:0/16:0) \\
%   %\hline  
%  \textbf{Cluster 2} 
%             Cer(d18:1/22:0)   GPCho(18:2/0:0) GPCho(16:0/20:5)  GPCho(18:0/20:4) \\  GPCho(18:2/20:4)       GPCho(18:0/22:6)   GPCho(16:0e/20:4)  \\GPCho(38:4e)  GPCho(18:1e/20:4) \\
% \textbf{Cluster 3}
% GPCho(18:3/0:0)
%      %  \hline
% \end{tabular}}\label{table:DIPPclusters}
% \end{table}
% {\begin{tabular}{@{}lll@{}}
%        \hline
%        \textbf{Cluster 1} & \textbf{Cluster 2} & \textbf{Cluster 3}\\
%        12~\% & 11.9~\% & 11.4~\% \\
%        \hline
%        TG(42:0) & Cer(d18:1/22:0) & GPCho(18:3/0:0)\\
%        TAG(16:0/14:0/16:0) & GPCho(18:2/0:0)\\
%        TAG(18:1/16:0/12:0) & GPCho(16:0/20:5) &\\
%        TG(46:2) & GPCho(18:0/20:4)\\
%        TAG(16:0/16:0/16:0) & GPCho(18:2/20:4)\\
%        TAG(18:1/14:0/16:0) & GPCho(18:0/22:6)\\
%        TAG(16:0/18:2/14:1) & GPCho(16:0e/20:4)\\
%        TAG(16:0/18:1/15:0) & GPCho(38:4e)\\
%        TAG(16:0/18:0/16:0) & GPCho(18:1e/20:4)\\
%        \hline
% \end{tabular}}\label{table:DIPPclusters}
% \end{table}
\section{Discussion}
We presented a novel method for translating biomarkers between
multiple species from multi-way, time-series experiments. The case we
addressed is when there are no \emph{a priori} known matching between the
variables in the two datasets, but only a similar experimental
design. The method estimates ANOVA-type multi-way covariate effects
for clusters of variables, and identifies and separates effects that
are shared between the species and effects that are species-specific.

For biological data, the task is harder than for generated
data. Probabilities of matched shared effects were lower for biological data, which is caused by the
fact that the covariate effects in a biological experiment are
weaker, making it more challenging for the method to find the
similarity between the data sets. The study with lipidomic T1D data
showed, however, that the method is able to extract similarities
between non-paired biological data sets. The approach presented can be naturally extended to multiple ways (covariates) and to multiple species. 
%\begin{thebibliography}{4}
%\bibliography{/share/mi/mi/doc/mi,/share/mi/mi/doc/gene/gene}
% \bibliography{/share/mi/mi/doc/sami,/share/mi/mi/doc/mi,/share/mi/mi/doc/gene/gene}
%\bibliography{../../doc/sami,../../doc/mi,../../doc/gene/gene}
%\begin{thebibliography}{4}
\bibliographystyle{splncs03}
\bibliography{../../doc/mi,../../doc/gene/gene}
%\end{thebibliography}
%\end{thebibliography}
\end{document}